\documentclass[pdflatex,sn-mathphys-num]{sn-jnl}% Math and Physical Sciences Numbered Reference Style

\usepackage{graphicx}%
\usepackage{multirow}%
\usepackage{amsmath,amssymb,amsfonts}%
\usepackage{amsthm}%
\usepackage{mathrsfs}%
\usepackage[title]{appendix}%
\usepackage{xcolor}%
\usepackage{textcomp}%
\usepackage{manyfoot}%
\usepackage{booktabs}%
\usepackage{algorithm}%
\usepackage{algorithmicx}%
\usepackage{algpseudocode}%
\usepackage{listings}%
\theoremstyle{thmstyleone}%
\theoremstyle{thmstyletwo}%

\theoremstyle{thmstylethree}%

\begin{document}

\title[Interactive 3D Cutaway Visualization]{Interactive 3D Cutaway Visualization for Subsurface Target Localization in Robot-Assisted Surgery}


% Authors
\author*[1]{\fnm{Yuxin} \sur{Chen}}
\email{yuxinchen@ece.ubc.ca}

\author[1]{\fnm{Wenxuan} \sur{Wang}}
\email{tiger001@student.ubc.ca}

\author[1]{\fnm{Wanwen} \sur{Chen}}
\email{wanwenc@ece.ubc.ca}

\author[1]{\fnm{Qi} \sur{Zeng}}
\email{qizeng@ece.ubc.ca}

\author[1]{\fnm{Septimiu E.} \sur{Salcudean}}
\email{tims@ece.ubc.ca}

% Affiliation
\affil*[1]{
    \orgdiv{Department of Electrical and Computer Engineering},
    \orgname{The University of British Columbia},
    \orgaddress{
        \city{Vancouver},
        \postcode{V6T 1Z4},
        \state{BC},
        \country{Canada}
    }
}

% \author*[1,2]{\fnm{First} \sur{Author}}\email{iauthor@gmail.com}



% \affil*[1]{\orgdiv{Department}, \orgname{Organization}, \orgaddress{\street{Street}, \city{City}, \postcode{100190}, \state{State}, \country{Country}}}

% \affil[2]{\orgdiv{Department}, \orgname{Organization}, \orgaddress{\street{Street}, \city{City}, \postcode{10587}, \state{State}, \country{Country}}}

% \affil[3]{\orgdiv{Department}, \orgname{Organization}, \orgaddress{\street{Street}, \city{City}, \postcode{610101}, \state{State}, \country{Country}}}


% \abstract{The abstract serves both as a general introduction to the topic and as a brief, non-technical summary of the main results and their implications. Authors are advised to check the author instructions for the journal they are submitting to for word limits and if structural elements like subheadings, citations, or equations are permitted.}

%%================================%%
%% Sample for structured abstract %%
%%================================%%

\abstract{
\textbf{Purpose:} Blended augmented reality (AR) overlays can provide ambiguous depth cues in which subsurface anatomy floats above the tissue surface. We investigate whether an interactive 3D cutaway visualization can improve robot-assisted target localization.

\textbf{Methods:} We propose Virtual Box, an interactive 3D cutaway interface that allows users to independently manipulate the viewing perspective to inspect subsurface targets.
% a methodology summary of the Virtual Box would be helpful here
%and the robot-tracked instrument tip. 
In a controlled, counterbalanced within-subject tissue phantom study, 15 engineering students used Virtual Box and a depth-enhanced blended AR overlay to position a da Vinci instrument at virtual targets in a water tank. We used registered 3D transrectal ultrasound to measure 3D localization error and depth error. We also collected completion time and subjective ratings.

\textbf{Results:} Virtual Box reduced mean 3D error from 12.92 to 4.96 mm (61.6\% reduction) and depth error from 9.48 to 2.07 mm (78.2\% reduction; both $p<0.001$). Spatial understanding, depth perception, and confidence ratings were higher (all unadjusted $p<0.001$), but completion time increased from 21.64 to 32.58 s ($p=0.008$). Overall workload did not differ significantly.

\textbf{Conclusion:} Virtual Box improved localization accuracy and perceived spatial understanding in this controlled task, with additional completion time, supporting further evaluation for surgical navigation.
}




\maketitle

% ============================================================
% 1. INTRODUCTION
% ============================================================

\section{Introduction}\label{sec1}

Robot-assisted minimally invasive surgery provides magnified stereoscopic visualization and precise instrument control. However, 
%with few exceptions such as the Da Vinci Firefly\TM camera, 
endoscopic imaging primarily depicts tissue surfaces, offering limited information about subsurface anatomy. This limitation is relevant to robot-assisted laparoscopic prostatectomy, where identifying tumor boundaries and adjacent neurovascular structures is important for balancing oncological control with functional preservation. MRI supports surgical planning~\cite{martini2019imageguidance}, while registered MRI and transrectal ultrasound (US) provide complementary anatomical information during robotic guidance~\cite{samei2020partial}. Beyond accurate registration, effective navigation requires users to interpret the three-dimensional (3D) relationships between subsurface anatomy, tissue surfaces, and surgical instruments.

Existing systems communicate these relationships through 3D navigation environments or blended AR overlays. 
MRI, US, and robotic instrument information were integrated within a registered 3D scene~\cite{samei2020partial}. 
Markerless calibration for camera-aligned AR
%~\cite{kalia2020markerless} 
was developed for~\cite{samei2020partial} and evaluated in~\cite{kalia2021preclinical}.
Multimodal visualization and depth-perceived interaction were studied in~\cite{wen2017multimodality}.
% SORRY BUT THIS PAPER PRESENTS RESULTS WITHOUT ANY EXPERIMENTAL VALIDATION FROM USERS. MAY BE INTERESTING BUT I DON: TWANT TO CITE IT UNTIL IT IS PEER-REVIEWED 
%MIRAGE~\cite{nasiri2026mirage} combined stereo reconstruction, ultrasound, instrument tracking, and multiple visualization modes. 
Multimodal guidance is feasible, but its effectiveness also depends on how registered information is presented.

In blended AR overlays, virtual subsurface structures may appear to float above tissue or produce conflicting depth cues;
%~\cite{sielhorst2006depth}.
these perceptual challenges may be partially addressed by non-photorealistic rendering with surface-based occlusion cues~\cite{lerotic2007pqspace}.
% The perceptual reliability of surgical AR visualizations was investigated~\cite{hettig2018arvr}, as well as luminance as a depth cue~\cite{reissis2023luminance}. Together, these studies motivate visualization strategies that communicate anatomical depth beyond spatial alignment alone.
%
Alternative approaches reveal occluded anatomy through contextual visualization~\cite{bichlmeier2007contextual}, %channels~\cite{wimmer2008cutaway}, alternative virtual viewpoints~\cite{navab2007mirror}, 
and interactive volume sectioning~\cite{chu2018hybrid}. 
Instrument-to-anatomy distance~\cite{choi2016depth} and multiple projection directions~\cite{ribeiro2024intraparenchymal} may improve localization accuracy. 
Selective anatomical overlays and depth-dependent transparency have also been investigated; 
AR helps localization and is preferred over separate displays~\cite{ramalhinho2025displays,olthof2026feasibility}. 
These approaches provide mechanisms for inspecting subsurface anatomy, but evidence remains limited on their combined use in a cutaway scene for physical targeting with an instrument
%
%Display and interaction choices affect localization performance. 
~\cite{ramalhinho2023value}. 

\begin{figure}[t]
\centering
\includegraphics[width=\linewidth]{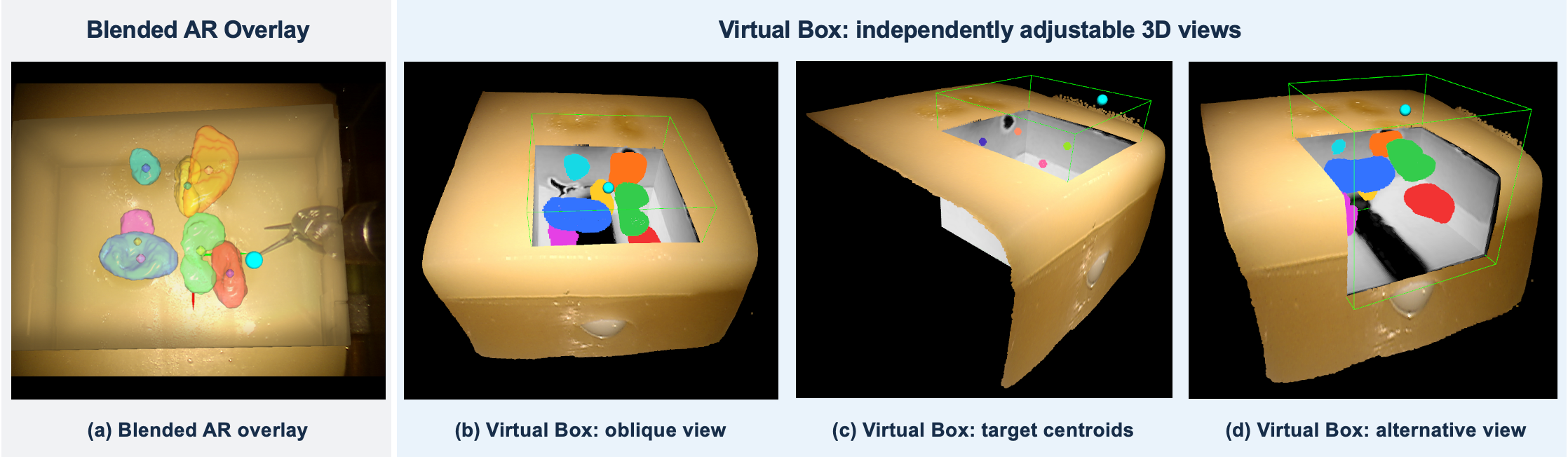}
\caption{Comparison of the visualization interfaces. (a) Depth-enhanced blended AR overlay on the live endoscopic image. (b--d) Virtual Box viewed from different perspectives, showing the reconstructed surface, localized cutaway, and registered cross-sectional images. Colored structures represent US-derived target segmentations. In (c), the target segmentations are hidden to reveal the small target-colored spheres marking target centroids, which define the desired instrument-tip positions. The bright cyan sphere with a dark outline indicates the robot-tracked instrument-tip position. The green wireframe delineates the inspection box.}
\label{fig:visualization_comparison}
\end{figure}

Prior work does not establish whether a manipulable cutaway scene improves physical robot-assisted subsurface localization relative to a depth-enhanced blended overlay.
To address this question, we propose Virtual Box, an interactive 3D scene that displays registered subsurface targets and the robot instrument tip. Users can position a cutaway box displaying dynamically updated MRI or US cross-sections and independently rotate the entire scene without moving the physical endoscope. Figure~\ref{fig:visualization_comparison} illustrates both visualization interfaces. A controlled within-subject study compares Virtual Box with a depth-enhanced blended AR overlay in terms of localization accuracy, completion time, and subjective assessments. The comparison evaluates the complete interfaces rather than isolating individual visualization features. Our main contributions are:

%\begin{enumerate}
{\em 1. An interactive Virtual Box interface} combining localized surface cutaway, dynamically updated multimodal cross-sections, independent viewpoint manipulation, and real-time robot-tracked instrument-tip visualization.

{\em 2. A controlled within-subject evaluation} on the da Vinci platform showing reduced 3D and depth localization errors and higher ratings for spatial understanding, depth perception, and confidence with Virtual Box, accompanied by longer completion times.
% \end{enumerate}

% ============================================================
% 2. METHODS
% ============================================================
\section{Methods}\label{sec:methods}

%\subsection{System Overview}\label{subsec:system}

We use a da Vinci Si Surgical System (Intuitive Surgical, USA) with a stereoscopic endoscope and a bk3500 US system (BK Medical) with an E14CL4B transducer mounted on a motorized rotational scanning mechanism. Stereo-reconstructed surface geometry, registered MRI and 3D US, and robot-tracked instrument positions supported two visualization conditions: Virtual Box and a depth-enhanced blended AR overlay, presented through TilePro at the console.

\subsection{Multimodal Reconstruction and Registration}\label{subsec:reconstruction}

Camera intrinsics, distortion parameters, and stereo extrinsics were calibrated using a ChArUco board and OpenCV. IGEV-Stereo~\cite{xu2023igev} estimated disparity from rectified stereo images, which was converted to metric depth using the calibrated focal length and baseline. Back-projection produced a colored surface point cloud from a reference stereo pair; this reconstruction remained fixed during localization trials.

Rotational US volumes were reconstructed from two-dimensional images using acquisition angles and calibrated probe geometry, following previous guidance systems~\cite{samei2020partial,kalia2021preclinical}. Targets were segmented using nnInteractive~\cite{isensee2025nninteractive} with manual refinement, and their centroids defined the reference target positions.

The fixed camera-to-ECM transformation was estimated using point-based calibration adapted from Kalia et al.~\cite{kalia2020markerless}. Robot-tracked 3D tip positions and manually annotated right-image locations were fitted using perspective-n-point estimation with RANSAC and nonlinear refinement. Corresponding US-measured and robot-tracked tip positions were aligned by RANSAC-based rigid registration~\cite{%adebar2011robotic
kalia2020markerless} to estimate $\mathbf{T}_{US\leftarrow DV}$, which maps positions from the robot reference frame ($DV$) into the US frame. Its inverse, $\mathbf{T}_{DV\leftarrow US}$, maps US-derived positions into the robot frame.

\begin{figure}[t]
\centering
\includegraphics[width=\linewidth]{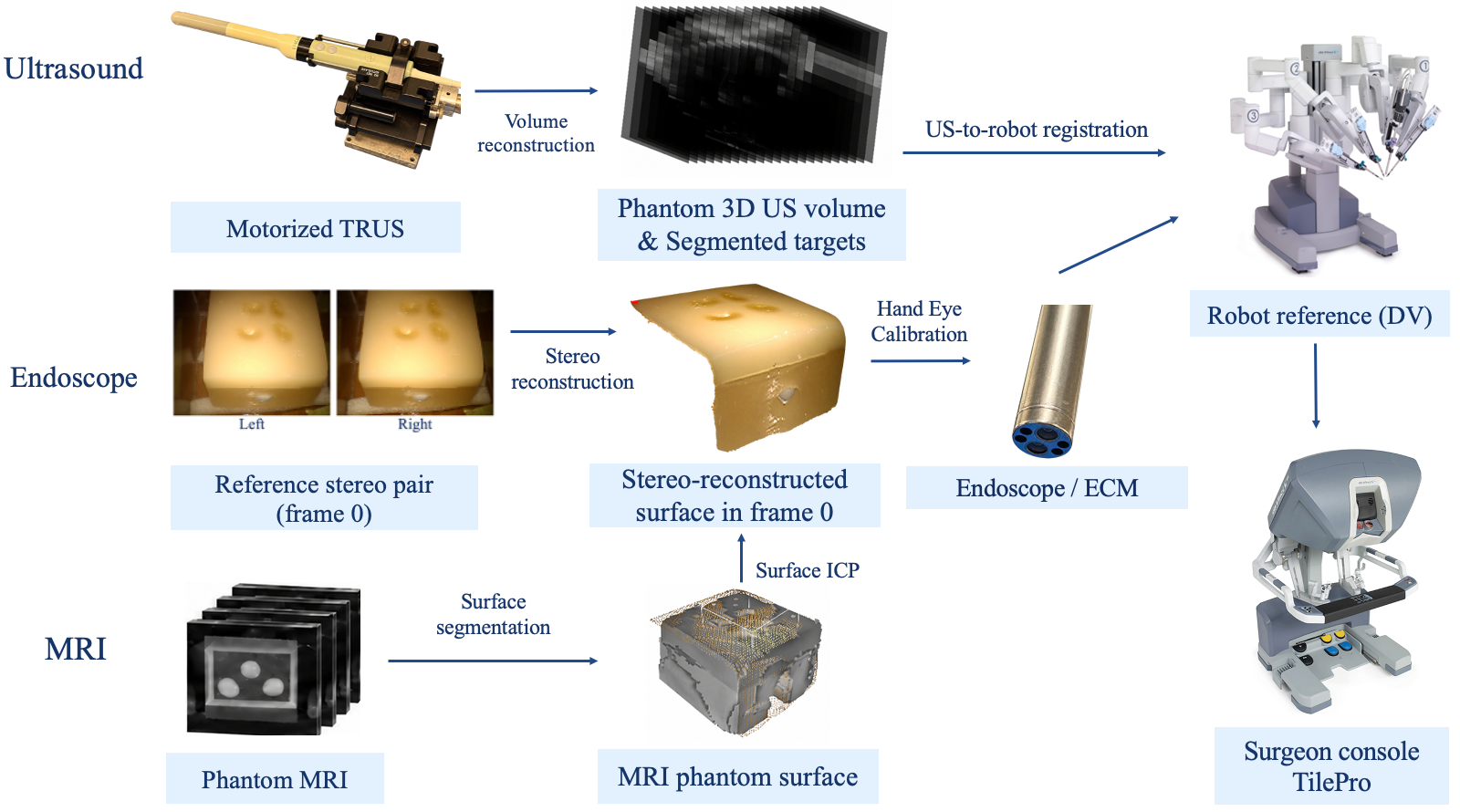}
\caption{Overview of multimodal reconstruction and registration. Rotational US provides the reconstructed volume and segmented targets. The reference stereo pair provides the camera-visible phantom surface, to which the MRI-derived phantom surface is registered using ICP. US-to-robot registration and camera-to-ECM calibration support integration of the imaging data and robot-tracked instrument-tip positions. Both visualization interfaces are presented through TilePro at the surgeon console; actual interface examples are shown in Fig.~\ref{fig:visualization_comparison}. MRI renderings are schematic.}
\label{fig:system_workflow}
\end{figure}

All data were expressed in the right-camera frame at the reference acquisition pose. With $\mathbf{T}_{A\leftarrow B}$ mapping coordinates from frame $B$ to frame $A$, the US-to-camera transformation was
$
\mathbf{T}_{C_0\leftarrow US}
=
\mathbf{T}_{C_0\leftarrow ECM}
\mathbf{T}_{ECM\leftarrow DV}
\mathbf{T}_{DV\leftarrow US},
\label{eq:coordinate_chain}
$
where $C_0$ and $ECM$ denote the reference right-camera and endoscopic camera manipulator frames, respectively. The ECM-dependent transformations were evaluated at the reference acquisition pose, and
$\mathbf{T}_{DV\leftarrow US}
=
\mathbf{T}_{US\leftarrow DV}^{-1}$.
Robot-tracked tip positions were mapped into the same camera frame using robot kinematics and camera-to-ECM calibration.
MRI was acquired from the same phantom used in the experiment. The MRI-derived surface was registered to the stereo-reconstructed surface using point-to-point ICP followed by point-to-plane refinement. The resulting transformation, $\mathbf{T}_{C_0\leftarrow MRI}$, was applied to the MRI volume. MRI and US were thus aligned indirectly through the common reference camera frame, rather than through direct MRI--US registration. Target segmentations were derived independently from US. Figure~\ref{fig:system_workflow} summarizes the multimodal reconstruction and registration workflow supporting both visualization interfaces.

\subsection{Interactive 3D Cutaway Visualization}\label{subsec:cutaway}

The proposed interactive 3D cutaway visualization presents the stereo-reconstructed surgical surface, registered MRI and US volumes, US-derived target segmentation, and robot-tracked instrument-tip position within a common 3D scene. Unlike blended AR overlays, the visualization allows users to manipulate both the anatomical inspection region and the viewing perspective independently of the physical endoscope.

A movable and rotatable inspection box defines a localized region of interest within the reconstructed scene. For a spatial point $\mathbf{p}$ expressed in the common coordinate system, its position relative to the inspection box is given by
\begin{equation}
\mathbf{q}_{B} = \mathbf{R}_{B}^{\top}(\mathbf{p}-\mathbf{c}_{B}), \qquad
\mathbf{p}\in B \iff |q_{B,i}|\leq h_i,\quad i\in\{x,y,z\},
\label{eq:cutaway_box}
\end{equation}
where $\mathbf{c}_{B}$ and $\mathbf{R}_{B}$ denote the box center and orientation, respectively, and $h_i$ represents the box half-extent along each local axis. Reconstructed surface points located inside the box are excluded from rendering, creating a localized cutaway while preserving the surrounding surface context. Conversely, the US-derived 3D target segmentation is displayed only within the box boundaries. Both clipping operations are updated as the box is translated or rotated.

The box faces provide cross-sectional visualization of the registered MRI and US volumes. Users can switch between MRI and US display modes while maintaining the same inspection-box pose. For each box face, image intensities are sampled from the selected volume at the corresponding spatial coordinates. As the box is translated or rotated, the cross-sectional images are dynamically resampled to reflect the updated position and orientation of the inspection region.\\
\indent
In addition to manipulating the inspection box, users can independently rotate the entire 3D scene to inspect the spatial relationships between the reconstructed surface, subsurface target, and instrument tip from different perspectives, including lateral views that expose target depth. During this visualization condition, the physical endoscope remains stationary. The instrument-tip position is continuously obtained from the da Vinci robot kinematics and represented by a sphere that updates in real time. Unlike the reconstructed surface and target segmentation, the tip indicator is not restricted by the inspection-box boundaries and remains visible when located outside the box. No explicit instrument-to-target distance feedback is provided. Participants directly controlled the inspection-box position and orientation and the virtual viewing perspective using a mouse.
\subsection{Depth-Enhanced Blended AR Overlay}\label{subsec:conventional_ar}
The AR condition used the same anatomical data and US-derived target coordinates as Virtual Box. Registered structures were transformed into the current right-camera frame using robot kinematics and camera-to-ECM calibration, projected onto the undistorted image using the corresponding intrinsics, and composited by alpha blending. Participants could reposition the physical endoscope, with the overlay updated using the current ECM pose.

Depth-dependent opacity followed
\begin{equation}
\alpha(z)
=
\alpha_{\min}
+
(\alpha_{\max}-\alpha_{\min})
\operatorname{clip}\left(
\frac{z-z_{\min}}{z_{\max}-z_{\min}},0,1
\right),
\label{eq:ar_depth_shading}
\end{equation}
where $z$ is camera-relative depth, $z_{\min}$ and $z_{\max}$ define the normalization range, and $\alpha_{\min}$ and $\alpha_{\max}$ define the opacity limits. Opacity increased with depth, motivated by previous surgical AR work~\cite{olthof2026feasibility}. The same mapping and fixed parameters were used across structures, participants, and trials.\\
\indent
The robot-tracked tip was displayed as a projected two-dimensional marker, independently of the opacity mapping. Figure~\ref{fig:visualization_comparison} compares the interfaces.

\subsection{Experimental Setup and System Validation}\label{subsec:experiment}

\begin{figure}[t]
\centering
\includegraphics[width=0.85\linewidth]{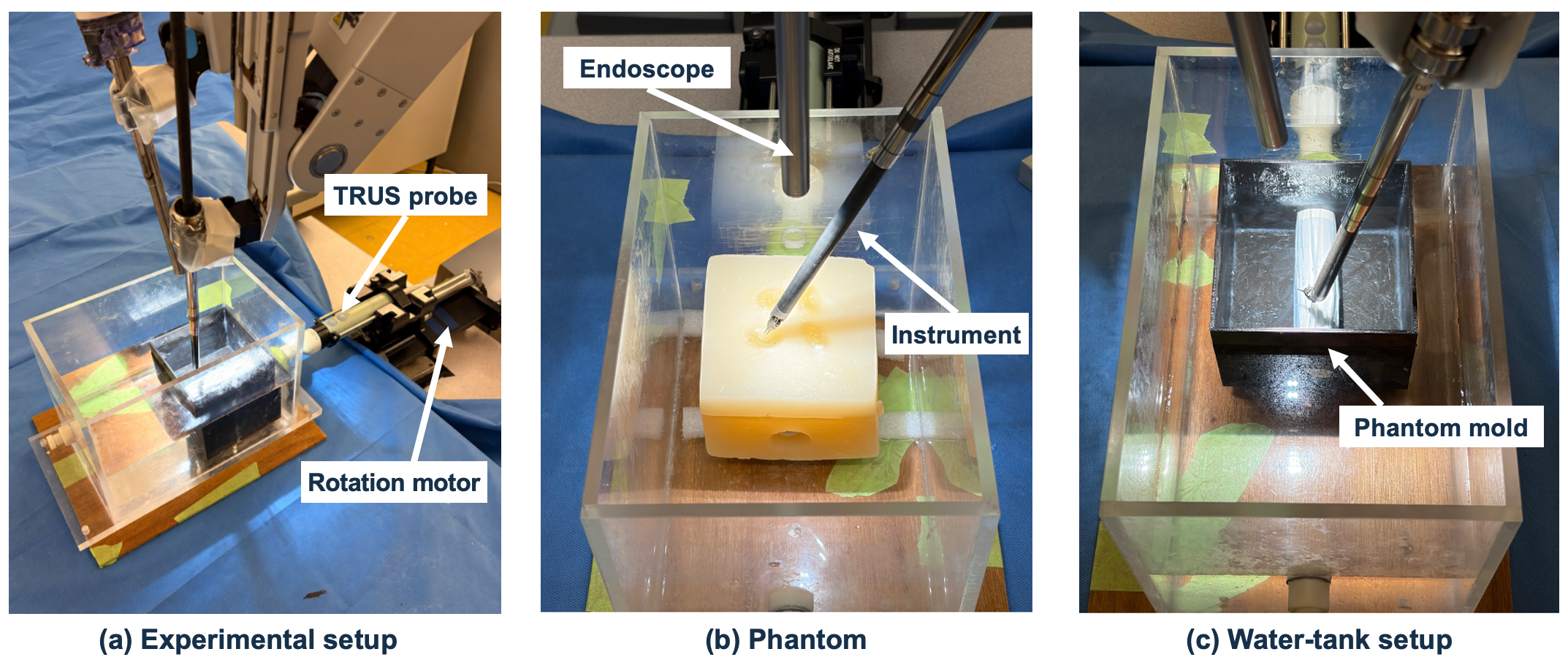}
\caption{Experimental apparatus for robot-assisted target localization. (a) Overall setup with robotic instrumentation and a motorized TRUS probe. (b) Gelatin phantom used to acquire the reference anatomy and target positions. (c) Water-tank setup used for instrument localization after phantom removal; the empty phantom mold provides a dark background.}
\label{fig:experimental_setup}
\end{figure}

A gelatin phantom containing fish-oil capsules provided six US-segmented targets: four shallow and two deep. Their centroids defined reference positions in the US frame, whose Z-axis represented depth. Stereo surface geometry, registered volumes, and target locations were retained as the reference scene. The phantom was then replaced by a water tank, permitting free instrument movement toward virtual targets without tissue penetration. The US transducer and positioning mechanism remained fixed to preserve the measurement reference. Figure~\ref{fig:experimental_setup} shows the apparatus. During targeting, Virtual Box retained the fixed surface reconstructed from the reference stereo pair, whereas the AR overlay was overlayed onto live endoscopic images of the water tank.

US-to-robot consistency was evaluated at 20 calibration-independent tip positions distributed across the workspace. Robot-tracked tip positions were transformed into the US frame and compared with US-measured positions. 3D Euclidean discrepancies and axis-wise RMS discrepancies were calculated in the US frame.

Camera projection alignment was evaluated using 20 calibration-independent tip samples across four ECM poses, with five samples per pose. Using the same calibration as the user study, robot-tracked tips were projected into the right image and compared with manually annotated physical tip locations. Reprojection error was the Euclidean pixel distance between predicted and annotated locations, summarized using mean, standard deviation, RMSE, and maximum error.

\subsection{User Study}\label{subsec:user_study}

A counterbalanced within-subject study compared the two interfaces. Each participant localized the same three randomly selected targets (two shallow and one deep) twice per condition, yielding 12 trials. Condition order was counterbalanced, and target order was randomized within each condition. Standardized instructions and practice trials preceded data collection.

Participants were instructed to stop when they judged the instrument tip to be positioned as accurately as possible, without numerical or color-coded distance feedback. Completion time ran from the start of instrument movement to confirmation of the final position. The instrument was then held stationary for rotational US imaging. The US-measured tip position and reference target were compared in the common US frame; robot-tracked positions were recorded separately for consistency assessment.

After completing both conditions, participants retrospectively evaluated each interface using two separate sets of measures. Seven agreement items assessed visualization quality (1 = strongly disagree; 7 = strongly agree); higher scores indicated more favorable assessments except for visual clutter. Participants were instructed that ``movable sphere'' referred to the instrument-tip position indicator.
Workload was assessed independently using the six NASA-TLX subscales: mental demand, physical demand, temporal demand, performance, effort, and frustration (0--100). \\
Overall Raw NASA-TLX was their unweighted mean and was not calculated from the seven agreement ratings. Higher TLX scores indicated greater workload; Performance ranged from 0 (perfect) to 100 (failure). An overall comparison assessed preference, perceived localization accuracy, depth judgment, and interference with the surgical view, with optional free-text feedback.

\subsection{Evaluation Metrics and Statistical Analysis}\label{subsec:evaluation}

The primary evaluation metric was 3D localization error, defined as the Euclidean distance between the US-measured final instrument-tip position and target centroid:
\begin{equation}
e_{s,c,j,r}^{3D}
=
\left\|
\mathbf{p}_{\mathrm{tip},s,c,j,r}^{US}
-
\mathbf{p}_{\mathrm{target},s,c,j,r}^{US}
\right\|_2,
\label{eq:localization_error}
\end{equation}
where $s$, $c$, $j$, and $r$ denote participant, condition, target, and repetition. The secondary evaluation metric was depth localization error, defined as the absolute coordinate difference between the tip and target along the US Z-axis.

These metrics quantify overall 3D positioning accuracy and accuracy along the experimental depth axis, respectively; lower errors indicate better localization. Tip and target positions were recorded as voxel indices in the reconstructed US volume, with isotropic voxel spacing of 0.5 mm. Coordinate differences were multiplied by 0.5 mm to calculate localization errors in millimeters.

Errors and completion times were averaged across six trials per participant and condition, yielding 15 paired observations per outcome. Shapiro--Wilk tests assessed normality of paired differences. 3D and depth errors were compared using two-sided paired $t$-tests, with mean paired differences and 95\% confidence intervals reported. Exploratory shallow and deep analyses used the same tests on participant-level means across four shallow or two deep trials per condition.

Completion time was exploratory and analyzed using a two-sided Wilcoxon signed-rank test because its paired differences departed from normality. The seven subjective ratings and overall Raw NASA-TLX scores were also analyzed using two-sided Wilcoxon signed-rank tests. Exact Wilcoxon $p$-values were obtained by enumerating sign assignments to nonzero paired differences, accounting for tied ranks. Questionnaire analyses were exploratory. All $p$-values were unadjusted for multiple comparisons, with $p<0.05$ considered statistically significant.

\section{Results}\label{sec:results}

%\subsection{System Accuracy Validation}
%\label{subsec:results_system_validation}

\begin{figure}[t]
\centering
\includegraphics[width=\linewidth]{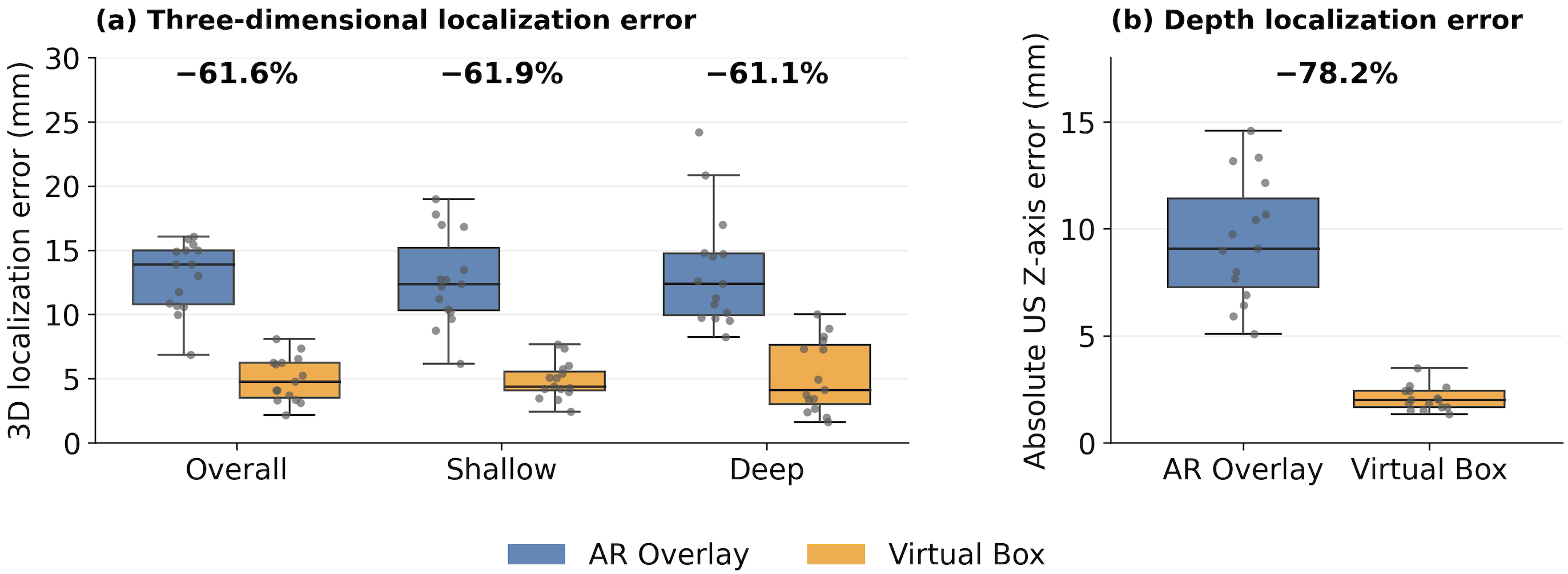}
\caption{Localization errors across 15 participants: (a) overall, shallow-, and deep-target 3D errors; (b) absolute US Z-axis depth error. Dots show participant means; boxes show medians and interquartile ranges. Percentages indicate mean error reductions with Virtual Box relative to blended AR.}
\label{fig:localization_results}
\end{figure}

{\bf System Accuracy Validation.} Across 20 calibration-independent positions, the 3D RMS discrepancy between US-measured and robot-tracked instrument-tip positions was 2.09 mm, with a maximum of 2.72 mm. RMS discrepancies along the US X-, Y-, and Z-axes were 0.89, 1.04, and 1.59 mm, respectively.
Across 20 independent tip samples acquired at four ECM poses using the user-study calibration, mean reprojection error was $25.57 \pm 10.40$ pixels, with an RMSE of 27.51 pixels and a maximum of 40.00 pixels in the $1280 \times 1024$ undistorted right-camera images.\\
%
%\subsection{Participant Characteristics}
%\label{subsec:participant_characteristics}
{\bf Participant Characteristics.} Fifteen engineering students completed all 180 trials. All had previous da Vinci experience; 13 reported experience with each of AR, US, and 3D visualization or image-guided navigation. Eight participants used the AR overlay first, and seven used Virtual Box first.\\
%
%\subsection{Three-Dimensional Target Localization Accuracy}
%\label{subsec:results_localization}
%
{\bf Three-Dimensional Target Localization Accuracy.} Virtual Box reduced mean 3D localization error from $12.92 \pm 2.69$ mm with the AR overlay to $4.96 \pm 1.74$ mm, a 61.6\% reduction. The mean paired difference was 7.96 mm (95\% CI: 6.27--9.65 mm; paired $t$-test, $p<0.001$); all 15 participants had lower mean errors with Virtual Box.\\
Exploratory participant-level paired $t$-tests also showed lower errors for shallow and deep targets (both $p<0.001$), with reductions of 61.9\% and 61.1\%, respectively. Table~\ref{tab:localization_results} reports the subgroup results, and Fig.~\ref{fig:localization_results} shows participant-level distributions.
%
%\subsection{Depth Localization Error and Task Completion Time}
%\label{subsec:results_depth_time}
%
{\bf Depth Localization Error and Task Completion Time.} Mean absolute depth error decreased from $9.48 \pm 2.90$ mm with the AR overlay to $2.07 \pm 0.57$ mm with Virtual Box, a 78.2\% reduction. The mean paired difference was 7.41 mm (95\% CI: 5.85--8.97 mm; paired $t$-test, $p<0.001$).
Mean completion time increased from $21.64 \pm 11.17$ s to $32.58 \pm 11.61$ s, an additional 10.94 s with Virtual Box (exploratory Wilcoxon signed-rank test, $p=0.008$).\\
%
%
%\subsection{Subjective Evaluation}
%\label{subsec:results_subjective}
%
{\bf Subjective Evaluation.} Participants rated Virtual Box higher in spatial understanding (median: 6 versus 3), depth perception (6 versus 2), and confidence (6 versus 3). Exploratory Wilcoxon signed-rank tests yielded unadjusted $p$-values of 0.000122, 0.000122, and 0.000244, respectively. No statistically significant differences were detected in visual clutter, target visibility, ease of interpretation, or visual comfort.
NASA-TLX scores can be seen in Table~\ref{tab:subjective_results}. Mean overall Raw NASA-TLX scores were 29.33 for Virtual Box and 34.78 for the AR overlay, with no statistically significant difference (Wilcoxon signed-rank test, $p=0.397$).
Overall, 14 participants (93.3\%) preferred Virtual Box, all 15 perceived it as providing more accurate localization, and 14 reported easier depth judgment. For reduced interference with the surgical view, five preferred Virtual Box, three preferred the AR overlay, and seven reported no difference.

\begin{table*}[t]
\caption{Localization errors for the two visualization conditions across 15 participants. Values are reported as mean $\pm$ standard deviation (mm), calculated from participant-level averages.}
\label{tab:localization_results}
\centering
\begin{tabular}{lccc}
\hline
Target depth & AR Overlay & Virtual Box & Reduction (\%) \\
\hline
Shallow & 12.70 $\pm$ 3.61 & 4.84 $\pm$ 1.43 & 61.9 \\
Deep & 13.36 $\pm$ 4.48 & 5.20 $\pm$ 2.81 & 61.1 \\
\hline
Overall & 12.92 $\pm$ 2.69 & 4.96 $\pm$ 1.74 & 61.6 \\
\hline
\end{tabular}
\end{table*}

\begin{table*}[t]
\caption{Subjective evaluation of 15 participants: median visualization ratings (1--7) and mean NASA-TLX scores (0--100). Higher ratings are better except for clutter; higher TLX scores are worse. Overall TLX averages six subscales.}
\label{tab:subjective_results}
\centering
\begin{tabular*}{\textwidth}{@{\extracolsep{\fill}}lcc@{}}
\hline
Measure & AR Overlay & Virtual Box \\
\hline
\multicolumn{3}{l}{\textit{Visualization ratings (1--7)}} \\
Spatial understanding & 3 & 6 \\
Depth perception & 2 & 6 \\
Visual clutter & 2 & 3 \\
Target visibility & 6 & 6 \\
Confidence & 3 & 6 \\
Ease of interpretation & 5 & 6 \\
Visual comfort & 6 & 6 \\
\hline
\multicolumn{3}{l}{\textit{NASA-TLX subscales (0--100)}} \\
Mental demand & 38.00 & 36.67 \\
Physical demand & 22.67 & 30.00 \\
Temporal demand & 29.33 & 30.00 \\
Performance & 51.33 & 26.00 \\
Effort & 34.67 & 38.67 \\
Frustration & 32.67 & 14.67 \\
\hline
Overall Raw NASA-TLX & 34.78 & 29.33 \\
\hline
\end{tabular*}
\end{table*}

\section{Discussion}\label{sec:discussion}

Virtual Box improved physical subsurface target localization compared with the depth-enhanced blended AR overlay in this controlled study. Mean 3D and depth errors decreased by 61.6\% and 78.2\%, respectively, accompanied by higher ratings for spatial understanding, depth perception, and confidence. Exploratory analyses showed improvements in both depth groups, although the limited target configurations preclude conclusions about depth-dependent benefits.

Blended overlays preserve image correspondence but may retain ambiguous depth and occlusion cues despite depth-dependent transparency. Virtual Box combines registered cross-sections and independent scene rotation to inspect target depth and instrument position from different directions without moving the endoscope. The results suggest improved spatial interpretation, but this complete-interface comparison cannot isolate the contributions of cutaway rendering, viewpoint manipulation, cross-sectional imaging, or tip representation. Localized cutaway and interactive sectioning have been explored previously, but recent advances in tissue-surface tracking under occlusion, instrument tracking and removal, and segmentation techniques that help initialize registration using tissue surfaces motivate revisiting this approach.

The additional 10.94 s per trial may reflect the effort required to manipulate and interpret the 3D scene, suggesting an accuracy--efficiency trade-off. Alternative viewpoints may also encourage further refinement, whereas ambiguous depth cues in blended AR may prompt earlier stopping. Nevertheless, 14 of 15 participants preferred Virtual Box, and no statistically significant difference in overall workload was detected. Future work should improve interaction efficiency and test these explanations using interaction logs and time-resolved localization errors.

The study has several limitations. Targets were derived from a tissue-mimicking phantom, but instrument positioning occurred in a water tank without tissue penetration, deformation, respiratory motion, or dynamic anatomical changes. Participants were engineering students rather than experienced surgeons, each evaluating three targets from a limited candidate set. Residual camera projection misalignment may also have affected the AR condition; its contribution could not be separated from interface design. Both interfaces were monoscopic; stereoscopic presentation warrants future evaluation. Secondary and exploratory findings should be interpreted with consideration of the unadjusted multiple comparisons. Finally, the remaining mean 3D error of 4.96 mm does not by itself establish clinical suitability. Future work will investigate dynamic reconstruction and deformable registration, improve interaction efficiency, and evaluate more clinically representative surgical tasks.

\section{Conclusion}\label{sec:conclusion}
Virtual Box integrates localized surface cutaway, registered MRI and US cross-sections, independent viewpoint manipulation, and robot-tracked instrument-tip visualization. 
In a controlled within-subject study involving 15 engineering students, it reduced mean 3D and depth localization errors by 61.6\% and 78.2\%, respectively, compared with a depth-enhanced blended AR overlay. Participants reported higher spatial understanding, depth perception, and confidence, although mean completion time increased by 10.94 s. These findings demonstrate an accuracy--efficiency trade-off between the evaluated interfaces during physical instrument localization to phantom-derived virtual targets in a water tank. Further evaluation is needed to establish applicability to clinically representative surgical tasks.

\backmatter

\section*{Declarations}

\noindent\textbf{Funding.}
This work was supported by a scholarship held by Y. Chen, NSERC and the C.A. Laszlo Chair held by Professor Salcudean.

\noindent\textbf{Competing interests.}
The authors declare no competing interests.

\noindent\textbf{Ethics approval and consent to participate.}
Approved by the UBC Research Ethics Board (H24-01947). Informed consent was obtained from all participants.

\noindent\textbf{Data and Code availability.}
Available from the authors upon reasonable request.

\bibliography{sn-bibliography}% common bib file

@inproceedings{xu2023igev,
  title={Iterative Geometry Encoding Volume for Stereo Matching},
  author={Xu, Gangwei and Wang, Xianqi and Ding, Xiaohuan and Yang, Xin},
  booktitle={Proceedings of the IEEE/CVF Conference on Computer Vision and Pattern Recognition},
  pages={21919--21928},
  year={2023}
}

@article{ramalhinho2023value,
  title={The value of Augmented Reality in surgery:
  A usability study on laparoscopic liver surgery},
  author={Ramalhinho, Joao and Yoo, Soojeong and
  Dowrick, Thomas and Koo, Bongjin and
  Somasundaram, Murali and Gurusamy, Kurinchi and
  Hawkes, David J. and Davidson, Brian and
  Blandford, Ann and Clarkson, Matthew J.},
  journal={Medical Image Analysis},
  volume={90},
  pages={102943},
  year={2023},
  doi={10.1016/j.media.2023.102943}
}

@article{kalia2020markerless,
  title={Evaluation of a marker-less, intra-operative,
  augmented reality guidance system for robot-assisted
  laparoscopic radical prostatectomy},
  author={Kalia, Megha and Mathur, Prateek and
  Tsang, Keith and Black, Peter and
  Navab, Nassir and Salcudean, Septimiu},
  journal={International Journal of Computer Assisted
  Radiology and Surgery},
  volume={15},
  number={7},
  pages={1225--1233},
  year={2020},
  doi={10.1007/s11548-020-02181-4}
}

@article{kalia2021preclinical,
  title={Preclinical evaluation of a markerless,
  real-time, augmented reality guidance system
  for robot-assisted radical prostatectomy},
  author={Kalia, Megha and Avinash, Apeksha and
  Navab, Nassir and Salcudean, Septimiu},
  journal={International Journal of Computer Assisted
  Radiology and Surgery},
  volume={16},
  number={7},
  pages={1181--1188},
  year={2021},
  doi={10.1007/s11548-021-02419-9}
}

@article{samei2020partial,
  title={A partial augmented reality system with
  live ultrasound and registered preoperative
  MRI for guiding robot-assisted radical
  prostatectomy},
  author={Samei, Golnoosh and Tsang, Keith and
  Kesch, Claudia and Lobo, Julio and Hor, Soheil
  and Mohareri, Omid and Chang, Silvia and
  Goldenberg, S. Larry and Black, Peter C.
  and Salcudean, Septimiu},
  journal={Medical Image Analysis},
  volume={60},
  pages={101588},
  year={2020},
  doi={10.1016/j.media.2019.101588}
}

@article{choi2016depth,
  author = {Choi, Hyunseok and
            Cho, Byunghyun and
            Masamune, Ken and
            Hashizume, Makoto and
            Hong, Jaesung},
  title = {An Effective Visualization Technique
           for Depth Perception in Augmented
           Reality-Based Surgical Navigation},
  journal = {The International Journal of
             Medical Robotics and Computer
             Assisted Surgery},
  volume = {12},
  number = {1},
  pages = {62--72},
  year = {2016},
  doi = {10.1002/rcs.1657}
}

@article{ramalhinho2025displays,
  author={Ramalhinho, Jo{\~a}o and
          Bulathsinhala, Sandun and
          Gurusamy, Kurinchi and
          Davidson, Brian R. and
          Clarkson, Matthew J.},
  title={Assessing augmented reality displays
         in laparoscopic liver surgery
         -- a clinical experience},
  journal={Surgical Endoscopy},
  volume={39},
  pages={5863--5871},
  year={2025},
  doi={10.1007/s00464-025-12008-5}
}

@misc{isensee2025nninteractive,
  author = {Isensee, Fabian and
            Rokuss, Maximilian and
            Kr{\"a}mer, Lars and
            Dinkelacker, Stefan and
            Ravindran, Ashis and
            Stritzke, Florian and
            Hamm, Benjamin and
            Wald, Tassilo and
            Langenberg, Moritz and
            Ulrich, Constantin and
            Deissler, Jonathan and
            Floca, Ralf and
            Maier-Hein, Klaus},
  title = {{nnInteractive}: Redefining 3D Promptable Segmentation},
  year = {2025},
  eprint = {2503.08373},
  archivePrefix = {arXiv},
  primaryClass = {cs.CV},
  doi = {10.48550/arXiv.2503.08373},
  note = {arXiv preprint}
}

@article{martini2019imageguidance,
  title   = {Image guidance in robot-assisted radical
             prostatectomy: how far do we stand?},
  author  = {Martini, Alberto and
             Wagaskar, Vinayak G. and
             Dell'Oglio, Paolo and
             Rastinehad, Ardeshir R. and
             Sfakianos, John P. and
             Tewari, Ashutosh K.},
  journal = {Current Opinion in Urology},
  volume  = {29},
  number  = {1},
  pages   = {10--13},
  year    = {2019},
  doi     = {10.1097/MOU.0000000000000565}
}

@article{wen2017multimodality,
  title   = {Augmented Reality Guidance with
             Multimodality Imaging Data and
             Depth-Perceived Interaction for
             Robot-Assisted Surgery},
  author  = {Wen, Rong and Chng, Chin-Boon and Chui, Chee-Kong},
  journal = {Robotics},
  volume  = {6},
  number  = {2},
  pages   = {13},
  year    = {2017},
  doi     = {10.3390/robotics6020013}
}

@inproceedings{lerotic2007pqspace,
  title     = {pq-space Based Non-Photorealistic Rendering
               for Augmented Reality},
  author    = {Lerotic, Mirna and Chung, Adrian J.
               and Mylonas, George and Yang, Guang-Zhong},
  booktitle = {Medical Image Computing and
               Computer-Assisted Intervention -- MICCAI 2007},
  series    = {Lecture Notes in Computer Science},
  volume    = {4792},
  pages     = {102--109},
  year      = {2007},
  publisher = {Springer},
  address   = {Berlin, Heidelberg}
}

@article{chu2018hybrid,
  title   = {Perception Enhancement Using Importance-Driven
             Hybrid Rendering for Augmented Reality Based
             Endoscopic Surgical Navigation},
  author  = {Chu, Yakui and Li, Xu and Yang, Xilin
             and Ai, Danni and Huang, Yong and Song, Hong
             and Jiang, Yurong and Wang, Yongtian
             and Chen, Xiaohong and Yang, Jian},
  journal = {Biomedical Optics Express},
  volume  = {9},
  number  = {11},
  pages   = {5205--5226},
  year    = {2018},
  doi     = {10.1364/BOE.9.005205}
}

@article{olthof2026feasibility,
  title   = {Toward Augmented Reality in Laparoscopic
             Liver Surgery Using Electromagnetic Tracking:
             A Clinical Feasibility Study},
  author  = {Olthof, Karin A. and Venix, Lisanne P. J.
             and Pruijt, Maaike and den Hartog, Anne G.
             and Kok, Niels F. M. and Ruers, Theo J. M.
             and Fusaglia, Matteo and Kuhlmann, Koert F. D.},
  journal = {Surgical Endoscopy},
  volume  = {40},
  pages   = {3538--3547},
  year    = {2026},
  doi     = {10.1007/s00464-026-12741-5}
}

@article{ribeiro2024intraparenchymal,
  title   = {Augmented Reality Guided Laparoscopic
             Liver Resection: A Phantom Study With
             Intraparenchymal Tumors},
  author  = {Ribeiro, Mathieu and
             Espinel, Yamid and
             Rabbani, Navid and
             Pereira, Bruno and
             Bartoli, Adrien and
             Buc, Emmanuel},
  journal = {Journal of Surgical Research},
  volume  = {296},
  pages   = {612--620},
  year    = {2024},
  doi     = {10.1016/j.jss.2023.12.014}
}

@inproceedings{bichlmeier2007contextual,
  title = {Contextual Anatomic Mimesis:
           Hybrid In-Situ Visualization Method
           for Improving Multi-Sensory Depth
           Perception in Medical Augmented Reality},
  author = {Bichlmeier, Christoph and
            Wimmer, Felix and
            Heining, Sandro Michael and
            Navab, Nassir},
  booktitle = {Proceedings of the 6th IEEE/ACM
               International Symposium on
               Mixed and Augmented Reality},
  pages = {129--138},
  year = {2007},
  doi = {10.1109/ISMAR.2007.4538837}
}

\end{document}